\documentclass{llncs}
\usepackage{a4,a4wide}
\usepackage{graphicx}
\usepackage{amssymb}
\usepackage{makeidx}  
\begin{document}
\mainmatter 
\title{Domain-specific Languages in a Finite Domain Constraint
  Programming System}
%
%
\author{Markus Triska}
\authorrunning{Markus Triska} 
%
%
\institute{Vienna University of Technology, Vienna, Austria,\\
\email{triska@dbai.tuwien.ac.at},\\ WWW home page:
\texttt{http://www.logic.at/prolog/}}

\maketitle              

\begin{abstract}
  In this paper, we present domain-specific languages~(DSLs) that we
  devised for their use in the implementation of a finite domain
  constraint programming system, available as \texttt{library(clpfd)}
  in SWI-Prolog and~YAP-Prolog. These DSLs are used in propagator
  selection and constraint reification. In these areas, they lead to
  concise specifications that are easy to read and reason about. At
  compilation time, these specifications are translated to Prolog
  code, reducing interpretative run-time overheads. The devised
  languages can be used in the implementation of other finite domain
  constraint solvers as well and may contribute to their correctness,
  conciseness and efficiency. 
\end{abstract}
\section{Introduction}
Domain-specific languages (DSLs) are languages tailored to a specific
application domain. DSLs are typically devised with the goal of
increased expressiveness and ease of use compared to general-purpose
programming languages in their domains of application (\cite{mernik}).
Examples of DSLs include \textit{lex} and \textit{yacc} (\cite{lex})
for lexical analysis and parsing, regular expressions for pattern
matching, HTML for document mark-up, VHDL for electronic hardware
descriptions and many other well-known instances.

DSLs are also known as ``\textit{little languages}''~(\cite{little}),
where ``little'' primarily refers to the typically limited intended or
main practical application scope of the language. For example,
PostScript is a ``little language'' for page descriptions.

CLP(FD), constraint logic programming over finite domains, is a
declarative formalism for describing combinatorial problems such as
scheduling, planning and allocation tasks~(\cite{jaffar}). It is one
of the most widely used instances of the general~CLP($\cdot$) scheme
that extends logic programming to reason over specialized domains.
Since~CLP(FD) is applied in many industrial settings like systems
verification, it is natural to ask: How can we implement constraint
solvers that are more reliable and more concise (i.e., easier to read
and verify) while retaining their efficiency? In the following
chapters, we present little languages that we devised towards this
purpose. They are already being used in a constraint solver over
finite domains, available as \texttt{library(clpfd)} in SWI-Prolog and
YAP-Prolog, and can be used in other systems as well.

\section{Related work}
In the context of CLP(FD), \textit{indexicals}~(\cite{diazclpfd}) are
a well-known example of a~DSL. The main idea of indexicals is to
declaratively describe the domains of variables as functions of the
domains of related variables. The indexical language consisting of the
constraint~``\texttt{in}'' and expressions such as
\texttt{min(X)..max(X)} also includes specialized constructs that make
it applicable to describe a large variety of arithmetic and
combinatorial constraints. GNU~Prolog~(\cite{gnuprolog}) and SICStus
Prolog~(\cite{openeneded}) are well-known Prolog systems that use
indexicals in the implementation of their finite domain constraint
solvers.

The usefulness of deriving large portions of code automatically from
shorter descriptions also motivates the use of \textit{variable
  views}, a DSL to automatically derive \textit{perfect} propagator
variants, in the implementation of Gecode~(\cite{gecode}).

\textit{Action rules}~(\cite{actionrules}) and Constraint Handling
Rules~(\cite{chr}) are Turing-complete languages that are very
well-suited for implementing constraint propagators and even entire
constraint systems (for example, B-Prolog's finite domain solver).

These examples of~DSLs are mainly used for the description and
generation of constraint \textit{propagation} code in practice. In the
following chapters, we contribute to these uses of~DSLs in the context
of CLP(FD) systems by presenting~DSLs that allow you to concisely
express selection of propagators and constraint reification with
desirable properties.

\section{Matching propagators to constraint expressions}
\label{sec:matching}

To motivate the DSL that we now present, consider the
following quote from Neng-Fa Zhou, author of B-Prolog~(\cite{bcsp}):

\begin{quote}
  A closer look reveals the reason [for failing to solve the problems
  within the time limit]: Almost all of the failed instances contain
  non-linear (e.g., $X*Y = C$, $abs(X-Y) = C$, and $X \hbox{mod}\ Y =
  C$) and disjunctive constraints which were not efficiently
  implemented in the submitted version of the solver.
\end{quote}

Consider the specific example of~$abs(X-Y) = C$: It is clear that
instead of decomposing the constraint into~$X-Y = T$, $abs(T) = C$, a
specialized combined propagator can be implemented and applied,
avoiding auxiliary variables and intermediate propagation steps to
improve efficiency. It is then left to detect that such a specialized
propagator can actually be applied to a given constraint expression.
This is the task of \textit{matching} available propagators to given
constraint expressions, or equivalently, mapping constraint
expressions to propagators.

Manually selecting fitting propagators for given constraint
expressions is quite error-prone, and one has to be careful not to
accidentally unify variables that occur in the expression with
subexpressions that one wants to check for. To simplify this task, we
devised a DSL in the form of a simple committed-choice language. The
language is a list of rules of the form~$M \rightarrow As$, where~$M$
is a matcher and~$As$ is a list of actions that are performed when~$M$
matches a posted constraint.

More formally, a \textit{matcher}~$M$ consists of the term~$m\_c(P,
C)$. $P$~denotes a \textit{pattern} involving a constraint
\textit{relation} like~\texttt{\#=}, \texttt{\#$>$} etc. and its
arguments, and~$C$ is a~\textit{condition} (a Prolog goal) that must
hold for a rule to apply. The basic building-blocks of a pattern are
explained in Table~\ref{margs}. These building-blocks can be nested
inside all symbolic expressions like addition, multiplication etc. A
rule is applicable if a given constraint is matched by~$P$ (meaning it
unifies with~$P$ taking the conditions induced by~$P$ into account),
and additionally~$C$ is true. A matcher~$m\_c(P, \mathtt{true})$, can
be more compactly written as~$m(P)$.

\begin{table}[ht]
  \centering
  \begin{tabular}[ht]{|c|l|}
    \hline
    \texttt{any(X)} & Matches any subexpression, unifying $X$ with that expression.\\
    \texttt{var(X)} & Matches a variable or integer, unifying~$X$ with it.\\
    \texttt{integer(X)} & Matches an integer, unifying~$X$ with it.\\
    \hline
  \end{tabular}
  \caption{Basic building-blocks of a pattern}
\label{margs}
\end{table}

In a rule~$M\rightarrow As$, each action~$A_i$ in the list of
actions~$As = [A_1,\dots,A_n]$ is one of the actions described in
Table~\ref{mactions}. When a rule is applicable, its actions are
performed in the order they occur in the list, and no further rules
are tried.

\begin{table}[ht]
  \centering
  \begin{tabular}[ht]{|c|p{10cm}|}
    \hline
    \texttt{g(G)} & Call the Prolog goal~$G$.\\
    \texttt{d(X, Y)} & Decompose arithmetic subexpression~$X$, unifying $Y$ with its result. Equivalent to \texttt{g(parse\_clpfd(X, Y))}, an internal predicate that is also generated from a similar DSL. \\
    \texttt{p(P)} & Post a constraint propagator~$P$. This is a shorthand notation for a specific sequence of goals that add a constraint to the constraint store and trigger it.\\
    \texttt{r(X, Y)} & Rematch the rule's constraint relation, using arguments~$X$ and~$Y$. Equi\-valent to \texttt{g(call(F,X,Y))}, where~$F$ is the functor of the rule's pattern.\\
    \hline
  \end{tabular}
  \caption{Valid actions in a list~$As$ of a rule $M\rightarrow As$}
\label{mactions}
\end{table}

\begin{figure}[ht]
  \centering
  \includegraphics[scale=0.9]{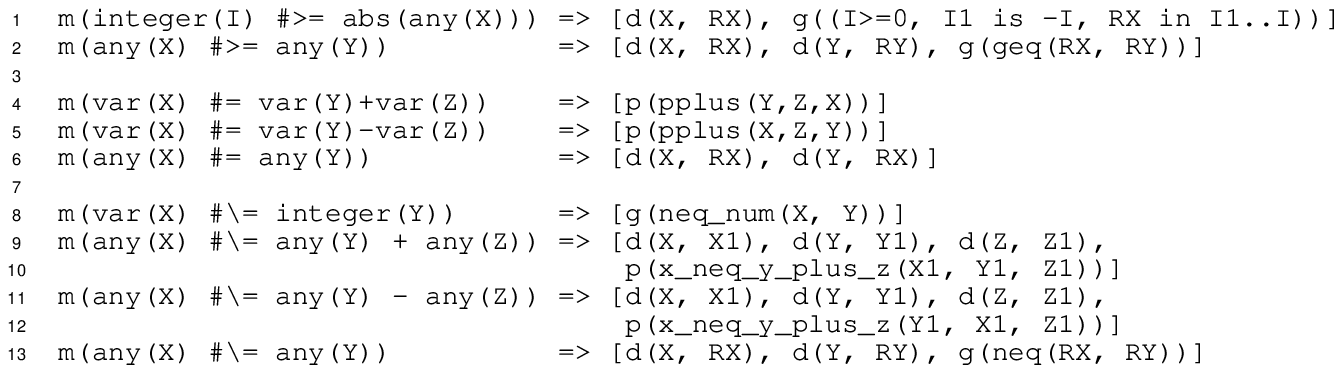}
  \caption{Rules for matching propagators in our constraint system. (Excerpt) }\label{fig:matchers}
\end{figure}

Figure~\ref{fig:matchers} shows some of the matching rules that we use
in our constraint system. It is only an excerpt; for example, in the actual system, nested additions are also detected and handled by a dedicated propagator. 
Such a declarative description has several advantages: First, it
allows automated subsumption checks to detect whether specialized
propagators are accidentally overshadowed by other rules. This is also
a mistake that we found easy to make and hard to detect when manually
selecting propagators. Second, when DSLs similar to the one we propose
here are also used in other constraint systems, it is easier to
compare supported specialized propagators, and to support common ones
more uniformly across systems. Third, improvements to the expansion
phase of the~DSL benefits potentially many propagators at
once. 

We found that the languages features we introduced above for matchers
and actions enable matching a large variety of intended specialized
propagators in practice, and believe that other constraint systems may
benefit from this or similar syntax as well.

\section{Constraint reification}
\label{sec:reification}
We now present a DSL that simplifies the implementation of constraint
\textit{reification}, which means reflecting the truth values of
constraint relations into Boolean $0/1$-variables.

When implementing constraint reification, it is tempting to proceed as
follows: For concreteness, consider reified equality~(\texttt{\#=/2})
of two CLP(FD)~expressions $A$ and~$B$. We could introduce two
temporary variables, $T_A$ and~$T_B$, and post the constraints
\texttt{$T_A$ \#= A} and \texttt{$T_B$ \#= B}, thus using the
constraint solver itself to decompose the (possibly compound)
expressions~$A$ and~$B$, and reducing reified equality of two
\textit{expressions} to equality of two finite domain
\textit{variables} (or integers), which is easier to implement.
Unfortunately, this strategy yields wrong results in general. Consider
for example the constraint (\texttt{\#<==>/2} denotes Boolean
equivalence):
$$\texttt{(X/0 \#= Y/0) \#<==> B}$$

It is clear that the relation~\texttt{X/0 \#= Y/0} cannot hold, since a
divisor can never be~0. A valid (declaratively equivalent) answer to
the above constraint is thus (note that~$X$ and~$Y$ must be
constrained to integers for the relation to hold):
$$\texttt{B = 0, X in inf..sup, Y in inf..sup}$$

However, if we decompose the equality \texttt{X/0 \#= Y/0} into two
auxiliary constraints \texttt{$T_A$ \#= X/0} and~\texttt{$T_B$ \#= Y/0}
and post them, then (with strong enough propagation of division)
both auxiliary constraints fail, and thus the whole query
(incorrectly) fails. While devising a DSL for reification, we found
one commercial Prolog system and one freely available system that
indeed incorrectly failed in this case. After we reported the issue,
the problem was immediately fixed.

It is thus necessary to take \textit{definedness} into account when
reifying constraints. See also~\cite{stuckey}, where our constraint
system (in contrast to others that were tested) correctly handles all
reification test cases, which we attribute in part to the DSL
presented in this chapter. Once any subexpression of a relation
becomes undefined, the relation cannot hold and its associated truth
value must be~$0$. Undefinedness can occur when~$Y=0$ in the
expressions $X/Y$, $X \hbox{mod}\ Y$, and $X \hbox{rem}\ Y$. Parsing
an arithmetic expression that occurs as an argument of a constraint
that is being reified is thus at least a ternary relation, involving
the expression itself, its arithmetic result, and its Boolean
definedness.

There is a fourth desirable component in addition to those just
mentioned: It is useful to keep track of \textit{auxiliary variables}
that are introduced when decomposing subexpressions of a constraint
that is being reified. The reason for this is that the truth value of
a reified constraint may turn out to be irrelevant, for instance the
implication~\texttt{0 \#==> $C$} holds for both possible truth values
of the constraint~$C$, thus auxiliary variables that were introduced
to hold the results of subexpressions while parsing~$C$ can be
eliminated. However, we need to be careful: A constraint propagator
may \textit{alias} user-specified variables with auxiliary variables.
For example, in \texttt{abs(X) \#= T, X \#>= 0}, a constraint system
may deduce~\texttt{X = T}. Thus, if~$T$ was previously introduced as
an auxiliary variable, and~$X$ was user-specified, $X$ must still
retain its status as a constrained variable.

\begin{figure}[ht]
  \centering
  \includegraphics{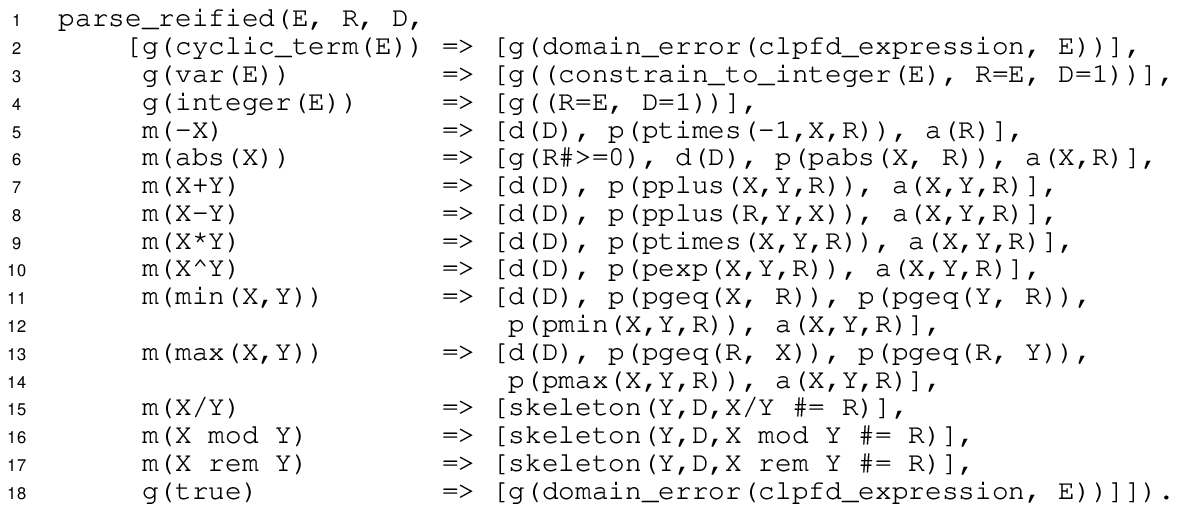}
  \caption{Parsing arithmetic expressions in reified constraints with our DSL}
\label{fig:reification}
\end{figure}

These considerations motivate the following DSL for parsing arithmetic
expressions in reified constraints, which we believe can be useful in
other constraint systems as well: A parsing rule is of the
form~$H\rightarrow Bs$. A head~$H$ is either a term~$g(G)$, meaning
that the Prolog goal~$G$ is true, or a term~$m(P)$, where~$P$ is a
symbolic pattern and means that the expression~$E$ that is to be
parsed can be decomposed as stated, recursively using the parsing
rules themselves for subterms of~$E$ that are subsumed by variables
in~$P$. The body~$Bs$ of a parsing rule is a list of body elements,
which are described in Table~\ref{tab:reification}. The predicate
\texttt{parse\_reified/4}, shown in Figure~\ref{fig:reification},
contains our full declarative specification for parsing arithmetic
expressions in reified constraints, relating an arithmetic
expression~$E$ to its result~$R$, Boolean definedness~$D$, and
auxiliary variables according to the given parsing rules, which are
applied in the order specified, committing to the first rule whose
head matches. This specification is again translated to Prolog code at
compile time and used in other predicates.

\begin{table}[ht]
  \centering
  \begin{tabular}[ht]{|c|p{8cm}|}
    \hline
    \texttt{g(G)} & Call the Prolog goal~$G$. \\
    \texttt{d(D)} & $D$ is 1 if and only if all subexpressions of~$E$ are defined. \\
    \texttt{p(P)} & Add the constraint propagator~$P$  to the constraint store. \\
    \texttt{a(A)} & $A$ is an auxiliary variable that was introduced while parsing the given compound expression~$E$. \\
    \texttt{a(X,A)} & $A$ is an auxiliary variable, unless \texttt{A == X}. \\
    \texttt{a(X,Y,A)} & $A$ is an auxiliary variable, unless \texttt{A == X} or \texttt{A == Y}. \\
    \texttt{skeleton(Y,D,G)} & A ``skeleton'' propagator is posted. When $Y$ cannot become~$0$ any more, it calls the Prolog goal~$G$  and binds~$D=1$. When~$Y$ is~0, it binds~$D=0$. When $D=1$ (i.e., the constraint must hold), it posts~\texttt{Y \#$\backslash$= 0}. \\
    \hline
  \end{tabular}
  \caption{Valid body elements for a parsing rule}
  \label{tab:reification}
\end{table}

\section{Conclusion and future work}
\label{sec:conclusion}
We have presented DSLs that are used in the implementation of a finite
domain constraint programming system. They enable us to capture the
intended functionality with concise declarative specifications. We
believe that identical or similar DSLs are also useful in the
implementation of other constraint systems. In the future, we intend
to generate even more currently hand-written code automatically from
smaller declarative descriptions.

%
%


\begin{thebibliography}{5}
%

\bibitem {mernik}
Mernik, M., Heering, J., Sloane, A. M.:
When and how to develop domain-specific languages.
ACM Comput. Surv., 37(4), 316--344 (2005)

\bibitem {lex} Johnson, S. C., Lesk, M. E.:
Language development tools.
Bell System Technical Journal, 56(6), 2155--2176 (1987)

\bibitem {little}
Bentley, J.: Little languages.
Communications of the ACM, 29(8), 711--21 (1986)

\bibitem {diazclpfd}
Codognet, P., Diaz, D.: Compiling Constraints in clp(FD).
Journal of Logic Programming, 27(3) (1996)

\bibitem {jaffar}
Jaffar, J., Lassez, J-L.: Constraint Logic Programming.
POPL, 111--119 (1987)

\bibitem {openeneded}
Carlsson, M., Ottosson, G., Carlson, B.: An Open-Ended Finite Domain Constraint Solver.
Proc. Prog. Lang.: Implementations, Logics, and Programs (1997)

\bibitem {gnuprolog}
Diaz, D., Codognet, P.: Design and Implementation of the GNU Prolog System.
Journal of Functional and Logic Programming (JFLP), Vol. 2001, No. 6 (2001)

\bibitem {gecode}
Schulte, Ch., Tack, G.: Perfect Derived Propagators. CoRR entry (2008)

\bibitem {actionrules}
Zhou, N-F.: Programming Finite-Domain Constraint Propagators in Action Rules.
Theory and Practice of Logic Programming, Vol.6, No.5, pp. 483--508 (2006)

\bibitem{chr}
Fr\"uhwirth, T.: Theory and Practice of Constraint Handling Rules.
Special Issue on Constraint Logic Programming, J. of Logic Programming, Vol 37(1--3) (1998)

\bibitem {bcsp}
Zhou, N-F.: A Report on the BPrologCSP Solver (2007)

\bibitem {stuckey}
Frisch, Alan M., Stuckey, Peter J.: The Proper Treatment of Undefinedness in Constraint Languages, CP 2009, LNCS 5732, 367--382 (2009)




  





\end{thebibliography}
\end{document}